%% file: cvpr.tex
\documentclass[final]{cvpr}

\usepackage{times}
\usepackage{epsfig}
\usepackage{graphicx}
\usepackage{amsmath}
\usepackage{amssymb}

\usepackage{times}
\usepackage{epsfig}
\usepackage{graphicx}
\usepackage{amsmath}
\usepackage{amssymb}
\usepackage{tabularx}
\usepackage{hhline}
\usepackage{subcaption}
\usepackage{blindtext}
\usepackage[breaklinks=true,bookmarks=false, hidelinks]{hyperref}

\makeatletter
\def\thickhline{%
	\noalign{\ifnum0=`}\fi\hrule \@height \thickarrayrulewidth \futurelet
	\reserved@a\@xthickhline}
\def\@xthickhline{\ifx\reserved@a\thickhline
	\vskip\doublerulesep
	\vskip-\thickarrayrulewidth
	\fi
	\ifnum0=`{\fi}}
\makeatother

\newcommand{\secrefede}[1]{Section \ref{#1}}

\newcommand{\eqnref}[1]{Eq.\ \ref{#1}}

\newcommand{\figrefede}[1]{Fig.\ \ref{#1}}
\newcommand{\figrefedetwo}[2]{Figs.\ \ref{#1} and \ref{#2}}
\newcommand{\figrefedethree}[3]{Figs.\ \ref{#1}, \ref{#2}, and \ref{#3}}
\newcommand{\tabrefede}[1]{Table \ref{#1}}
\newcommand{\tabrefedetwo}[2]{Tables \ref{#1} and \ref{#2}}

\newcommand{\norm}[1]{\left\lVert#1\right\rVert}

\newlength{\thickarrayrulewidth}
\def\cvprPaperID{8305} % *** Enter the CVPR Paper ID here
\def\confYear{CVPR 2021}
\begin{document}

%%%%%%%%% TITLE
\title{Back to Event Basics: Self-Supervised Learning of\\ Image Reconstruction for Event Cameras via Photometric Constancy}

\author{Federico Paredes-Vall\'es \qquad Guido C. H. E. de Croon\\
	Micro Air Vehicle Laboratory, Delft University of Technology, The Netherlands
	%\small \{f.paredesvalles, g.c.h.e.decroon\}@tudelft.nl
}

\maketitle

%%%%%%%%% ABSTRACT
\begin{abstract}
	Event cameras are novel vision sensors that sample, in an asynchronous fashion, brightness increments with low latency and high temporal resolution. The resulting streams of events are of high value by themselves, especially for high speed motion estimation. However, a growing body of work has also focused on the reconstruction of intensity frames from the events, as this allows bridging the gap with the existing literature on appearance- and frame-based computer vision. Recent work has mostly approached this problem using neural networks trained with synthetic, ground-truth data. In this work we approach, for the first time, the intensity reconstruction problem from a self-supervised learning perspective. Our method, which leverages the knowledge of the inner workings of event cameras, combines estimated optical flow and the event-based photometric constancy to train neural networks without the need for any ground-truth or synthetic data. Results across multiple datasets show that the performance of the proposed self-supervised approach is in line with the state-of-the-art. Additionally, we propose a novel, lightweight neural network for optical flow estimation that achieves high speed inference with only a minor drop in performance.
\end{abstract}
%\begin{center}
%	\vspace{-21pt}
%	\textbf{Supplementary material:}\\
%	\href{http://mavlab.tudelft.nl/ssl_e2v/}{http://mavlab.tudelft.nl/ssl\_e2v/}
%\end{center}

%%%%%%%%% BODY TEXT
\section{Introduction}

Unlike conventional cameras recording intensity frames at fixed time intervals, event cameras sample light based on scene dynamics by asynchronously measuring per-pixel brightness\footnote{Defined as the logarithm of the pixel intensity, i.e., $L\doteq \log(I)$.} changes at the time they occur \cite{gallego2019event}. This results in streams of sparse events encoding the polarity of the perceived changes. Because of this paradigm shift, event cameras offer several advantages over their frame-based counterparts, namely low power consumption, high dynamic range (HDR), low latency and high temporal resolution. %These properties make them very suitable for conditions that are challenging for standard cameras, e.g., HDR scenes and high speed motion.

Despite the advantages, the novel output format of event cameras poses new challenges in terms of algorithm design. Unless working with spiking neural networks \cite{paredes2020unsupervised}, events are usually converted into intermediate representations that facilitate the extraction of information \cite{gallego2019event}. Among others, intensity frames are an example of a powerful representation since they allow the evaluation of the appearance of a visual scene, thus bridging the gap between event cameras and the existing frame-based computer vision literature \cite{rebecq2019events, rebecq2019high}. For this reason, there has been a significant research drive to develop new methods to reconstruct images from events with similar statistics to those captured by standard cameras. 

\begin{figure}[!t]
	\centering
	\includegraphics[width=0.485
	\textwidth]{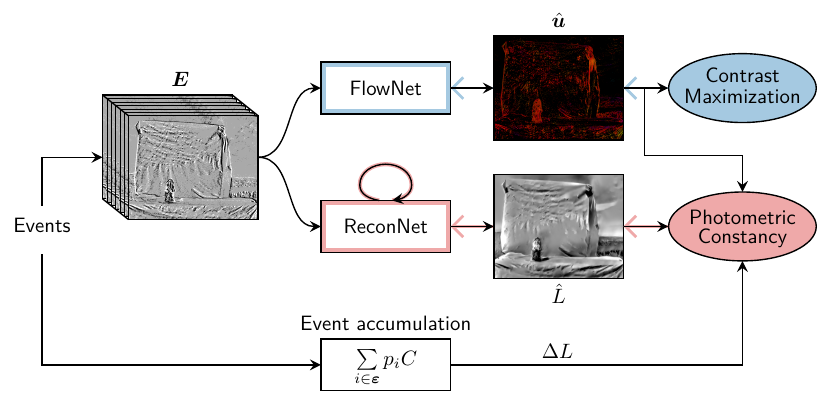}
	\caption{Overview of the proposed framework. Our model is trained in a self-supervised fashion to perform optical flow estimation and image reconstruction from event data using the contrast maximization proxy loss and the event-based photometric constancy, respectively. Colored reverse arrows indicate error propagation for each loss.}
	\label{fig:overview}
\end{figure}

Recent work has mostly approached this problem from a machine learning perspective. With their E2VID artificial neural network, Rebecq \textit{et al.}\ \cite{rebecq2019events, rebecq2019high} were the first to show that learning-based methods trained to maximize perceptual similarity via supervised learning outperform hand-crafted techniques by a large margin in terms of image quality. Later, Scheerlinck \textit{et al.}\ \cite{scheerlinck2020fast} achieved high speed inference with FireNet, a simplified model of E2VID. Despite the high levels of accuracy reported, these architectures were trained with large sets of synthetic data from event camera simulators \cite{rebecq2018esim}, which adds extra complexity to the reconstruction problem due to the \textit{simulator-to-reality gap}. In fact, Stoffregen, Scheerlinck \textit{et al.}\ \cite{stoffregen2020train} recently showed that if the statistics of the synthetic training datasets do not closely resemble those seen during inference, image quality degrades and the generalizability of these architectures remains limited.

In this work, we propose to come back to the theoretical basics of event cameras to relax the dependency of learning-based reconstruction methods on ground-truth and synthetic data. Specifically, we introduce the self-supervised learning (SSL) framework in \figrefede{fig:overview}, which consists of two artificial neural networks, \textit{FlowNet} and \textit{ReconNet}, for optical flow estimation and image reconstruction, respectively. FlowNet is trained through the contrast maximization proxy loss from Zhu \textit{et al.}\ \cite{zhu2019unsupervised}, while ReconNet makes use of the flow-intensity relation in the \textit{event-based photometric constancy} \cite{gallego2015event} to reconstruct the frames that best satisfy the input events and the estimated flow. Using our method, we retrain several networks from the image reconstruction \cite{rebecq2019events, scheerlinck2020fast} and optical flow \cite{zhu2018ev} literature. In terms of accuracy, results show that the reconstructed images are in line with those generated by most learning-based approaches despite the lack of ground-truth data during training. Additionally, we propose \textit{FireFlowNet}, a lightweight architecture for optical flow estimation that, inspired by \cite{scheerlinck2020fast}, achieves high speed inference with only a minor drop in performance.

In summary, this paper contains \textit{two main contributions}. First, a novel SSL framework to train artificial neural networks to perform event-based image reconstruction that, with the aid of optical flow, does not require ground truth of any kind and can learn directly on real event data. Second, we introduce FireFlowNet: a novel, lightweight neural network architecture that performs fast optical flow estimation from events. We validate our self-supervised method and optical flow network through extensive quantitative and qualitative evaluations on multiple datasets.

\section{Related Work}

Early methods to image reconstruction from event data approached the problem through the \textit{photometric constancy}: each event provides one equation relating the intensity gradient and the optical flow \cite{gallego2015event}. Kim \textit{et al.}\ \cite{kim2008simultaneous} were the first in the field and developed an Extended Kalman Filter that, under rotational and static scene assumptions, reconstructs a gradient image that is later transformed into the intensity space via Poisson integration. They later extended this approach to 6 degrees-of-freedom camera motion \cite{kim2016real}. Under the same assumptions, Cook \textit{et al.}\ \cite{cook2011interacting} simultaneously recovered intensity images, optical flow, and angular velocity through bio-inspired, interconnected network of interacting maps. Bardow \textit{et al.}\ \cite{bardow2016simultaneous} developed a variational energy minimization framework to simultaneously estimate optical flow and intensity from sliding windows of events, relaxing for the first time the static scene assumption.

Instead of relying on the photometric constancy, several approaches based on direct event integration have been proposed, which do not assume scene structure or motion dynamics. Reinbacher \textit{et al.}\ \cite{reinbacher2016real} formulated intensity reconstruction as an energy minimization problem via direct integration with periodic manifold regularization. Scheerlinck \textit{et al.}\ \cite{Scheerlinck18accv} achieved computationally efficient reconstruction by filtering events with a high-pass filter prior to integration.

%Inspired by the learning-based approach of Barua \textit{et al.}\ \cite{barua2016direct}, 

Several machine learning approaches have also been proposed. Training generative adversarial networks with real grayscale frames was proposed by Wang \textit{et al.}\ \cite{wang2019event} and Pini \textit{et al.}\ \cite{pini2019learn}. However, Rebecq \textit{et al.}\ \cite{rebecq2019events, rebecq2019high} showed that training in a supervised fashion with a large synthetic dataset allowed for higher quality reconstructions with their \textit{E2VID} architecture. Focused on computational efficiency, Scheerlinck \textit{et al.}\ \cite{scheerlinck2020fast} managed to significantly reduce E2VID complexity with \textit{FireNet}, with only a minor drop in accuracy. Inspired by these works, Choi \textit{et al.}\ \cite{choi2020learning} and Wang \textit{et al.}\ \cite{wang2020eventsr} recently proposed hybrid approaches that incorporate super resolution aspects in the training process and architecture design to improve image quality. Lastly, Stoffregen, Scheerlinck \textit{et al.}\ \cite{stoffregen2020train} recently highlighted that, when training with ground truth, the statistics of the training dataset play a major role in the reconstruction quality. They showed that a slight change in the training statistics of E2VID leads to significant improvements across multiple datasets. %; and their refined E2VID+ architecture currently is state-of-the-art. 

Our proposed SSL framework (see \figrefede{fig:overview}) is based on the event-based photometric constancy used by early reconstruction methods. Similarly to Bardow \textit{et al.}\ \cite{bardow2016simultaneous}, we simultaneously estimate intensity and optical flow from the input events. However, instead of relying on a joint optimization scheme, we achieve it via two independent neural networks that only share information during training. Further, we reconstruct intensity directly from the photometric constancy, instead of from an oversimplified model of the event camera. This approach allows, for the first time, to relax the strong dependency of learning-based approaches on ground-truth and synthetic data.%, with only a small drop in reconstruction quality.

\section{Method}\label{sec:method}

An event camera consist of an array of independent pixels that respond to changes in the brightness signal $L(t)$, and transmit these changes through streams of sparse and asynchronous events \cite{lichtsteiner2008128}. For an ideal camera, an event $\boldsymbol{e}_i=(\boldsymbol{x}_i,t_i,p_i)$ is triggered at pixel $\boldsymbol{x}_i=(x_i,y_i)^T$ and time $t_i$ whenever the brightness change since the last event at that pixel reaches a contrast sensitivity threshold $C$. Therefore, the brightness increment occurred in a time window $\Delta t_k$ is encoded in the event data via pixel-wise accumulation:
\begin{align}\label{eqn:deltaevents2}
\Delta L_k(\boldsymbol{x}) = \sum_{\boldsymbol{e}_i\in \Delta t_k} p_i C
\end{align}
where $C>0$, and the polarity $p_i\in\{+,-\}$ encodes the sign of the brightness change.

As in \cite{gallego2015event}, under the assumptions of Lambertian surfaces, constant illumination and small $\Delta t$, we can linearize \eqnref{eqn:deltaevents2} to obtain the event-based photometric constancy:
\begin{equation}\label{eqn:deltamodel}
\Delta L_k(\boldsymbol{x})\approx -\nabla L_{k-1}(\boldsymbol{x})\cdot \boldsymbol{u}_k(\boldsymbol{x})\Delta t_k
\end{equation}
which encodes that events are caused by the spatial gradients of the brightness signal, $\nabla L=(\delta_x L, \delta_y L)^T$, moving with optical flow $\boldsymbol{u}=(u, v)^T$. The dot product conveys that no events are generated if the flow vector is parallel to an edge ($\boldsymbol{u}\bot\nabla L$), while they are generated at the highest rate if perpendicular ($\boldsymbol{u}\parallel\nabla L$). Thus, events are caused by the projection of the optical flow vector in the $\nabla L$ direction.%, the so-called normal flow \cite{lucas1981iterative}. %As in \cite{lucas1981iterative}, we further refer to this projection as \textit{normal optical flow}.% and denote it by $\hat{\boldsymbol{u}}(\boldsymbol{x})$.

\subsection{Overview}

Our goal is to learn, in an SSL fashion, to transform a continuous stream of events into a sequence of intensity images $\smash{\{\hat{I}_k\}}$. To achieve this, we propose the pipeline in \figrefede{fig:overview} in which two neural networks are jointly trained. On the one hand, \textit{FlowNet} is a convolutional network that learns to estimate optical flow by compensating for the motion blur in the input events. On the other hand, \textit{ReconNet} is a recurrent convolutional network that learns to perform image reconstruction through the event-based photometric constancy. %), which, as shown in \figrefede{fig:overview}, makes use of FlowNet's estimates and the brightness increments obtained from event accumulation.% (\eqnref{eqn:deltaevents2}).%, where $\smash{\hat{\bar{I}}_k\in\left[0, 1\right]^{W\times H}}$

\begin{figure*}[!t]
	\centering
	\includegraphics[width=0.875
	\textwidth]{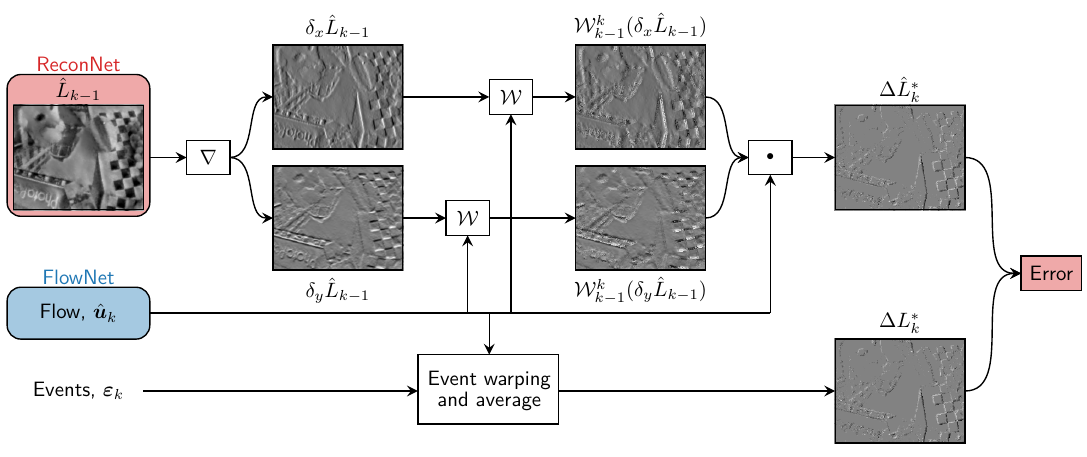}
	\caption{Brightness reconstruction via the event-based photometric constancy formulation proposed in this work. The most recent event-based optical flow estimate from FlowNet $\hat{\boldsymbol{u}}_k$ is used to (i) warp the input events, (ii) warp the spatial gradients of the last reconstructed image  $\hat{L}_{k-1}$, and (iii) in the dot product with the warped gradients. The predicted brightness increment image $\Delta \hat{L}_k^{*}$ is compared to that obtained with the deblurred (and averaged) input events, $\Delta L_k^{*}$, and the error is propagated backwards towards ReconNet to improve reconstruction accuracy.}
	\label{fig:constancy}
\end{figure*}

\subsection{Input Event Representation}\label{sec:voxel}

As proposed in \cite{zhu2019unsupervised}, the input to both our networks is a voxel grid $E_k$ with $B$ temporal bins that gets populated with consecutive, non-overlapping partitions of the event stream $\boldsymbol{\varepsilon}_k\doteq\{\boldsymbol{e}_i\}_{i=0}^{N-1}$, each containing a fixed number of events, $N$. For each partition, every event (with index $i$) distributes its polarity $p_i$ to the two closest bins according to:
\begin{align}
E(\boldsymbol{x}_i, t_b) &= \sum_i p_i \kappa(t_b-t_{i}^{*}\left(B-1\right))\label{eqn:inputone}\\
\kappa(a) &= \max(0, 1-|a|)\\
t_{i}^{*} &= \frac{\left(t_i-t_{0}^{k}\right)}{\left(t_{N-1}^{k}-t_{0}^{k}\right)}\label{eqn:inputtwo}
\end{align}
where $b$ is the bin index, and $t_{i}^{*}\in\left[0,1\right]$ denotes the normalized event timestamp. This representation adaptively normalizes the temporal dimension of the input depending on the timestamps of each partition of events.% $\boldsymbol{\varepsilon}_k$. %Therefore, the networks do not have access to absolute temporal information.

\subsection{Optical Flow via Contrast Maximization}\label{sec:flow}

We aim to learn to reconstruct $L$ through the photometric constancy in \eqnref{eqn:deltamodel}, which, besides the spatial and temporal derivatives of the brightness itself, also depends on the optical flow  $\boldsymbol{u}$. One could use ground-truth optical flow to solve for this ill-posed problem. However, due to the limited availability of event-camera datasets with accurate ground-truth data, we opt for training our FlowNet to perform flow estimation in a self-supervised manner, using the contrast maximization proxy loss for motion compensation \cite{gallego2018unifying}.

A partition of events is said to be blurry whenever there is a spatiotemporal misalignment among its events, i.e., events generated by the same portion of a moving edge are captured with different timestamps and pixel locations. The idea behind the motion compensation framework \cite{gallego2018unifying} is that accurate optical flow can be retrieved by finding the motion model of each event that best deblurs $\boldsymbol{\varepsilon}_k$. Knowing the per-pixel optical flow, the events can be propagated to a reference time $t_{\text{ref}}$ through:
\begin{align}\label{eqn:warp}
\boldsymbol{x}'_i = \boldsymbol{x}_i + (t_{\text{ref}} - t_i)\boldsymbol{u}(\boldsymbol{x}_i)
\end{align}

In this work, we adopt the deblurring quality measure proposed by Mitrokhin \textit{et al.}\ \cite{mitrokhin2018event} and later refined by Zhu \textit{et al.}\ \cite{zhu2019unsupervised}: the per-pixel and per-polarity average timestamp of the resulting image of warped events (IWE), $H$. The lower this metric, the better the deblurring. As in \cite{zhu2019unsupervised}, we generate an image of the average (normalized) timestamp at each pixel for each polarity $p'$ via bilinear interpolation:
\begin{align}\label{eqn:flowloss}
\begin{aligned}
T_{p'}(\boldsymbol{x}{;}\boldsymbol{u} |t_{\text{ref}}^{*}) &= \frac{\sum_{j} \kappa(x - x'_{j})\kappa(y - y'_{j})t_{j}^{*}}{\sum_{j} \kappa(x - x'_{j})\kappa(y - y'_{j})+\epsilon}\\
j = \{i \mid p_{i}=&p'\}, \hspace{15pt}p'\in\{+,-\}, \hspace{15pt} \epsilon\approx 0
\end{aligned}
\end{align}
and minimize the sum of the squared images resulting from warping the events forward and backward to prevent scaling issues during backpropagation:
\begin{align}
\mathcal{L}_{\text{contrast}}(t_{\text{ref}}^{*}) &= \sum_{\boldsymbol{x}}T_{+}(\boldsymbol{x}{;}\boldsymbol{u} |t_{\text{ref}}^{*})^2+T_{-}(\boldsymbol{x}{;}\boldsymbol{u} |t_{\text{ref}}^{*})^2\\
\mathcal{L}_{\text{contrast}} &= \mathcal{L}_{\text{contrast}}(1) + \mathcal{L}_{\text{contrast}}(0)
\end{align}

The total loss used to train FlowNet is then given by:
\begin{equation}\label{eqn:totalflow}
\mathcal{L}_{\text{FlowNet}} = \mathcal{L}_{\text{contrast}} + \lambda_1 \mathcal{L}_{\text{smooth}}
\end{equation}
where $\mathcal{L}_{\text{smooth}}$ is a Charbonnier smoothness prior \cite{charbonnier1994two}, and $\lambda_1$ is a scalar balancing the effect of the two losses. Note that, since $\mathcal{L}_{\text{contrast}}$ does no propagate the error back to pixels without events, we mask FlowNet's output so that null optical flow vectors are returned at these pixel locations.

\subsection{Reconstruction via Photometric Constancy}\label{sec:recons}

We formulate the SSL reconstruction problem from an image registration perspective \cite{lucas1981iterative} via brightness increment images. Specifically, we propose to use the difference between the reference increment image $\Delta L$ (event integration, \eqnref{eqn:deltaevents2}) and the predicted $\Delta \hat{L}$ (photometric constancy, \eqnref{eqn:deltamodel}) to reconstruct the brightness signal that best explains the input events, assuming known error-free optical flow. This reconstructed brightness is denoted by $\hat{L}$. FlowNet predictions are used in the computation of $\Delta \hat{L}$, and as registration parameters to warp both increment images to a common temporal frame (indicated by the superscript $\text{}^*$). A schematic of the proposed formulation is shown in \figrefede{fig:constancy}.

To minimize motion blur in the reconstructed frames, instead of directly integrating the input events, we define the reference brightness increment $\Delta L^{*}$ via the per-pixel and per-polarity average number of warped events:
\begin{align}
\Delta L^{*}(\boldsymbol{x}{;}\boldsymbol{u})\doteq C\left(G_{+}(\boldsymbol{x}{;}\boldsymbol{u}|1)-G_{-}(\boldsymbol{x}{;}\boldsymbol{u}|1)\right)\\
\begin{aligned}
G_{p'}(\boldsymbol{x}{;}\boldsymbol{u}|t_{\text{ref}}^{*}) &= \frac{H_{p'}(\boldsymbol{x}{;}\boldsymbol{\boldsymbol{u}}|t_{\text{ref}}^{*})}{P_{p'}(\boldsymbol{x}{;}\boldsymbol{\boldsymbol{u}}|t_{\text{ref}}^{*}) + \epsilon}\\
%	p'\in\{+&,-\}, \hspace{15pt} \epsilon\approx 0
\end{aligned}\hspace{25pt}
\end{align}
where $P$ is a two-channel image containing the number of pixel locations from where the IWE $H$ receives events in the event warping process (see \secrefede{sec:flow}). Therefore, $\Delta L^{*}$ is a deblurred representation of the contrast change encoded in the input events. An ablation study on the impact of event deblurring prior to event integration can be found in the supplementary material.

On the other hand, we adapt the event-based photometric constancy in \eqnref{eqn:deltamodel} and compute $\Delta \hat{L}$ by warping the spatial gradients of the last reconstructed image to the current time instance via spatial transformers \cite{jaderberg2015spatial}:
\begin{align}\label{eqn:pred}
\Delta \hat{L}^{*}(\boldsymbol{x}{;}\boldsymbol{u})\doteq -\mathcal{W}_{k-1}^{k}(\nabla \hat{L}_{k-1}(\boldsymbol{x}))\cdot \hat{\boldsymbol{u}}_k(\boldsymbol{x})%\Delta t_k
\end{align}
where $\mathcal{W}_{k-1}^{k}$ is the warping function of the optical flow $\hat{\boldsymbol{u}}_k$.

Following a maximum likelihood approach \cite{lichtsteiner2008128, gehrig2020eklt}, we define the photometric reconstruction loss as the squared $L_2$ norm of the difference of the warped brightness increments:
\begin{align}\label{eqn:reconloss}
\mathcal{L}_{\text{PE}} &= \norm{\Delta L^{*}(\boldsymbol{x}{;}\boldsymbol{u}) - \Delta \hat{L}^{*}(\boldsymbol{x}{;}\boldsymbol{u})}_{2}^{2}
\end{align}
where, besides $\hat{L}$, the contrast threshold $C$ is the only remaining unknown. To relax the dependency on this parameter, our ReconNet uses linear activation in its last layer instead of the frequently used sigmoid function \cite{rebecq2019high, scheerlinck2020fast}. The resulting unbounded brightness estimate is first transformed into the intensity space through $\hat{I}_k=\exp(\hat{L}_k)$, and then linearly normalized to get the final reconstruction $\smash{\hat{I}_k^f}$:
\begin{align}
\begin{aligned}
\hat{I}_k^f &= \frac{\hat{I}_k - m}{M - m}
\end{aligned}
\end{align}
where $m$ and $M$ are the $1\%$ and $99\%$ percentiles of $\hat{I}_k$, and $\hat{I}_k^f$ is clipped to the range $\left[0,1\right]$. This min/max normalization allows the use of any value of $C$ for training as long as the ratio of positive and negative contrast thresholds resembles that of the evaluation sequences. We assume that most event-camera datasets were recorded with $C_{+}/C_{-}\approx 1$, and set both thresholds to $1$.

On its own, \eqnref{eqn:reconloss} is not sufficient for the reconstruction of temporally consistent images. Because of the dot product in \eqnref{eqn:pred}, the absence of input events can be ambiguously understood as lack of apparent motion, lack of spatial image gradients, or both. To solve for this issue, we introduce an explicit temporal consistency loss based on the frame-based formulation of the photometric constancy \cite{jason2016back}. In essence, we define the temporal loss as the photometric error between two successive reconstructed frames:
\begin{align}
\begin{aligned}
\mathcal{L}_{\text{TC}} = &\norm{\hat{L}_{k} - \mathcal{W}_{k-1}^{k}(\hat{L}_{k-1})}_1%_{2}^{2}
\end{aligned}
\end{align}

The total loss used to train ReconNet is then given by:
\begin{align}
\mathcal{L}_{\text{ReconNet}} = \sum_{k=0}^{S} \mathcal{L}_{\text{PE}} + \lambda_2 \sum_{k=S_0}^{S} \mathcal{L}_{\text{TC}} + \lambda_3 \sum_{k=0}^{S} \mathcal{L}_{\text{TV}}
\end{align} 
where $S$ denotes the number of steps we unroll the recurrent network for during training, $\mathcal{L}_{\text{TV}}$ is a smoothness total-variation constraint \cite{rudin1992nonlinear}, and $\lambda_2$ and $\lambda_3$ are scalars balancing the effect of the three losses.

\begin{figure}[!t]
	\centering
	\includegraphics[width=0.475
	\textwidth]{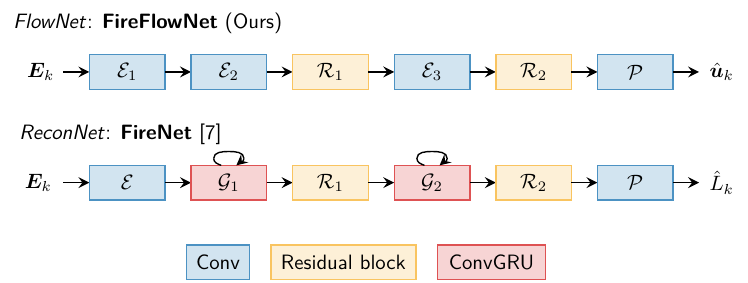}
	\vspace{-32.5pt}
	\begin{center}
		\line(1,0){240}
	\end{center}
	\vspace{-1pt}
	\includegraphics[width=0.475
	\textwidth]{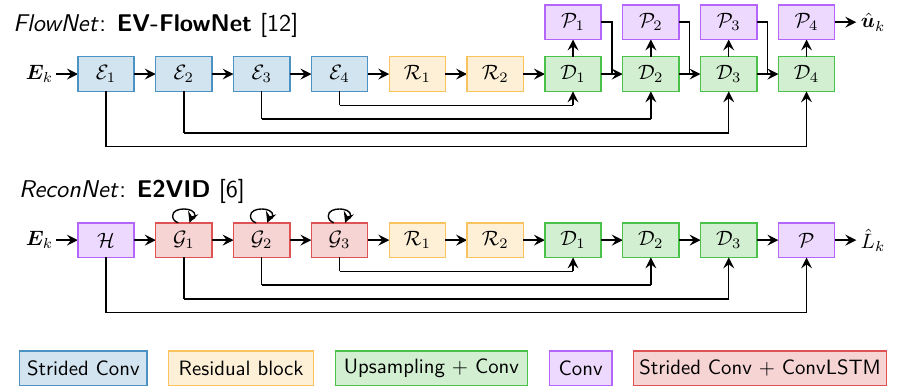}
	\caption{Neural networks evaluated in this work. %\textit{Top}: Single-strided networks. \textit{Bottom}: UNet-like \cite{ronneberger2015u} networks.
	}
	\label{fig:nets}
\end{figure}

\subsection{Network Architectures}

We evaluate the two trends on network design for event cameras when trained with our SSL framework. The evaluated architectures are shown in \figrefede{fig:nets}.

\noindent\textbf{FlowNet: FireFlowNet.} FireFlowNet is our proposed lightweight architecture for fast optical flow estimation. Inspired by FireNet \cite{scheerlinck2020fast}, the network consists of three encoder layers that perform single-strided convolutions, two residual blocks \cite{he2016deep}, and a final prediction layer that performs depthwise (i.e., $1\times 1$) convolutions with two output channels. All layers have 32 output channels and use $3\times 3$ kernels and ReLU activations except for the final, which uses tanh activations. A comparison of the key architectural differences between our FireFlowNet and the current state-of-the-art is shown in \tabrefede{tab:flownet}.

\noindent\textbf{FlowNet: EV-FlowNet \cite{zhu2018ev}.} The input voxel grid $\boldsymbol{E}_k$ is passed through four strided convolutional layers with output channels doubling after each layer starting from $64$. The resulting activations are then passed through two residual blocks \cite{he2016deep} and four decoder layers that perform bilinear upsampling followed by convolution. After each decoder, there is a (concatenated) skip connection from the corresponding encoder, as well as another depthwise convolution to produce a lower scale flow estimate, which is then concatenated with the activations of the previous decoder. The $\mathcal{L}_{\text{FlowNet}}$ loss (see \eqnref{eqn:totalflow}) is applied to each intermediate flow estimate via flow upsampling. All layers use $3\times 3$ convolutional kernels and ReLU activations except for the flow prediction layers, which use tanh activations. 

\begin{table}[!t]
	\caption{Main architectural differences between our FireFlowNet and EV-FlowNet \cite{zhu2018ev}. FireFlowNet has $250\times$ fewer parameters, consuming only $0.41\%$ of the memory.}
	\label{tab:flownet}
	\centering
	\resizebox{\linewidth}{!}{%
		\begin{tabular}{lcc}
			\thickhline
			& EV-FlowNet \cite{zhu2018ev} & FireFlowNet (Ours)\\\hline
			No. params. (k) & 14130.28 & 57.03 \\
			Memory (Mb) & 53.90 & 0.22 \\
			Downsampling   & Yes & No \\
			\thickhline
	\end{tabular}}
\end{table}

\begin{table*}[!t]
	\caption{Quantitative evaluation of our FlowNet architectures on the MVSEC dataset \cite{zhu2018multivehicle}. For each sequence, we report the AEE (lower is better, $\downarrow$) in pixels and the percentage of points with endpoint error greater than 3 pixels, $\%_{\text{Outlier}}$ ($\downarrow$). Best in bold, runner up underlined.
	}
	\label{tab:flowevluation}
	\centering
	\resizebox{0.85\linewidth}{!}{%
		{\renewcommand{\arraystretch}{1.1} 
			\begin{tabular}{lccccccccccc}
				\thickhline
				& \multicolumn{2}{c}{outdoor\_day1}&& \multicolumn{2}{c}{indoor\_flying1} && \multicolumn{2}{c}{indoor\_flying2} && \multicolumn{2}{c}{indoor\_flying3}\\\cline{2-3}\cline{5-6}\cline{8-9}\cline{11-12}
				& AEE & $\%_{\text{Outlier}}$&& AEE & $\%_{\text{Outlier}}$&& AEE & $\%_{\text{Outlier}}$&& AEE & $\%_{\text{Outlier}}$\\\hline
				EV-FlowNet$\text{}_{\text{GT-SIM}}$ \cite{stoffregen2020train} & 0.68 & 1.0 && \textbf{0.56} & \underline{1.0} && \textbf{0.66} & \textbf{1.0} && \textbf{0.59} & \textbf{1.0} \\
				EV-FlowNet$\text{}_{\text{FW-MVSEC}}$ \cite{zhu2018ev} & \underline{0.49} & \underline{0.2}&& 1.03 & 2.2 && 1.72 & 15.1 && 1.53 & 11.9 \\
				EV-FlowNet$\text{}_{\text{EW-MVSEC}}$ \cite{zhu2019unsupervised} & \textbf{0.32} & \textbf{0.0}&& \underline{0.58} & \textbf{0.0} && \underline{1.02} & \underline{4.0} && \underline{0.87} & \underline{3.0} \\
				EV-FlowNet$\text{}_{\text{EW-DR}}$ (Ours) & 0.92 & 5.4&& 0.79 & 1.2 && 1.40 & 10.9 && 1.18 & 7.4 \\
				FireFlowNet$\text{}_{\text{EW-DR}}$ (Ours) & 1.06 & 6.6 && 0.97 & 2.6 && 1.67 & 15.3 && 1.43 & 11.0 \\
				\thickhline
	\end{tabular}}}
\end{table*}

%\begin{table*}[!t]
%	\caption{Computational cost evaluation of our FireFlowNet architecture against EV-FlowNet \cite{zhu2018ev}. We report inference time on GPU and CPU, and the number of floating point operations (FLOPs) per forward-pass at common sensor resolutions. We used a single NVIDIA GeForce GTX 1080 Ti GPU and and Intel 2.60 GHz Xeon E5-2623 v4 CPU for all experiments.
%	}
%	\label{tab:flowmetrics}
%	\centering
%	\resizebox{0.85\linewidth}{!}{%
%		{\renewcommand{\arraystretch}{1.1} 
%			\begin{tabular}{lcccccccc}
%				\thickhline
%				& \multicolumn{2}{c}{GPU (ms)}&& \multicolumn{2}{c}{CPU (ms)} && \multicolumn{2}{c}{FLOPs (G)}\\\cline{2-3}\cline{5-6}\cline{8-9}
%				& EV-FlowNet & FireFlowNet && EV-FlowNet & FireFlowNet && EV-FlowNet & FireFlowNet \\\hline
%				$240\times 180$ & 4.33 & \textbf{1.97}&& 160.89 & \textbf{45.97} && 8.91 & \textbf{2.47} \\
%				$346\times 260$ & 7.05 & \textbf{3.81}&& 421.55 & \textbf{106.38} && 18.60 & \textbf{5.14} \\
%				$640\times 480$ & 17.04 & \textbf{12.55}&& 1243.95 & \textbf{397.59} && 61.47 & \textbf{17.59} \\
%				$1280\times 720$ & 49.32 & \textbf{34.24}&& 3458.84 & \textbf{1182.64} && 184.41 & \textbf{52.67} \\
%				\thickhline
%	\end{tabular}}}
%\end{table*}

\noindent\textbf{ReconNet: FireNet \cite{scheerlinck2020fast}.} Same architecture as FireFlowNet except for the second and third encoder, which are recurrent ConvGRU layers \cite{ballas2015delving}. As in \cite{scheerlinck2020fast}, each layer has 16 output channels, but we use linear activation in the final layer.

\noindent\textbf{ReconNet: E2VID \cite{rebecq2019high}.} The input voxel grid $\boldsymbol{E}_k$ is passed through a convolutional head layer, three recurrent encoders performing strided convolution followed by ConvLSTM \cite{xingjian2015convolutional}, two residual blocks \cite{he2016deep}, three decoder layers that perform bilinear upsampling followed by convolution, and a final depthwise convolutional prediction layer. There are (element-wise sum) skip connections between symmetric encoder and decoder layers, and the number of output channels in the head layer is 32 and doubles after each encoder. Head, encoder, and decoder layers use $5\times 5$ kernels, while the rest uses $3\times 3$. All layers use ReLU activations except for the final prediction layer which uses linear. %Contrary to the original E2VID, we do not use batch normalization \cite{ioffe2015batch}.

\section{Experiments}

We train our networks on the indoor forward facing sequences from the UZH-FPV Drone Racing Dataset (DR) \cite{delmerico2019we}, which is characterized by a much wider distribution of optical flow vectors than other datasets, such as MVSEC \cite{zhu2018multivehicle}, the Event-Camera Dataset (ECD) \cite{mueggler2017event}, or the High Quality Frames (HQF) dataset \cite{stoffregen2020train}. Our training sequences consist of approximately 15 minutes of event data recorded with a racing quadrotor flying aggressive  six-degree-of-freedom trajectories. We split these recordings and generate 440 $128\times 128$ (randomly cropped) sequences of 2 seconds each, and use them for training with $B=5$. We further augment this data using random horizontal, vertical and polarity flips, besides with artificial pauses of the input event stream (i.e., forward-pass with null input voxel). For training, we fixed the number of input events per pixel to $0.3$.

\begin{table}[!t]
	\caption{Computational cost evaluation of our FireFlowNet against EV-FlowNet \cite{zhu2018ev}. We report inference time on GPU and the floating point operations (FLOPs) per forward-pass at common sensor resolutions. We used a single NVIDIA GeForce GTX 1080 Ti GPU for all experiments.
	}
	\label{tab:flowmetrics}
	\centering
	\resizebox{\linewidth}{!}{%
		{\renewcommand{\arraystretch}{1.1} 
			\begin{tabular}{lccccc}
				\thickhline
				& \multicolumn{2}{c}{GPU (ms)}&& \multicolumn{2}{c}{FLOPs (G)}\\\cline{2-3}\cline{5-6}
				& EV-FlowNet & FireFlowNet && EV-FlowNet & FireFlowNet \\\hline
				$240\times 180$ & 4.33 & \textbf{1.97}&& 8.91 & \textbf{2.47} \\
				$346\times 260$ & 7.05 & \textbf{3.81}&& 18.60 & \textbf{5.14} \\
				$640\times 480$ & 17.04 & \textbf{12.55}&& 61.47 & \textbf{17.59} \\
				$1280\times 720$ & 49.32 & \textbf{34.24}&& 184.41 & \textbf{52.67} \\
				\thickhline
	\end{tabular}}}
\end{table}

Our framework is implemented in PyTorch\footnote{The project's code and additional qualitative results can be found at \href{http://mavlab.tudelft.nl/ssl_e2v/}{http://mavlab.tudelft.nl/ssl\_e2v/}.}. We use the Adam optimizer \cite{kingma2014adam} and a learning rate of $0.0001$ for both networks, and train with a batch size of 1 for 120 epochs. We empirically set the weights for each loss to $\{\lambda_1, \lambda_2, \lambda_3\} =\{1.0, 0.1, 0.05\}$, ReconNet's unrolling $S$ to 20 steps, and $S_0$ to $10$ steps.

\subsection{Optical Flow Evaluation}

To validate FireFlowNet as a lightweight alternative to the current state-of-the-art in event-based optical flow estimation, we evaluated both of our FlowNet architectures on the indoor\_flying and outdoor\_day sequences from the MVSEC dataset \cite{zhu2018multivehicle} with the ground-truth data provided by Zhu \textit{et al.}\ \cite{zhu2018ev}. Optical flow predictions were generated at each grayscale frame timestamp, and scaled to be the displacement between two successive frames. %For the ground-truth comparison, we convert the output of our network from units of pixels/grid into units of pixel displacement between frames as follows:
%\begin{align}
%	\hat{\boldsymbol{u}}_{\text{disp}} = \hat{\boldsymbol{u}}\times\frac{t_{F}^{k} - t_{F-1}^{k}}{t_{N-1}^{k} - t_{0}^{k}}
%\end{align}
%where the subscript $F$ is the index of the grayscale frame with respect to which the event partition $\boldsymbol{\varepsilon}_k$ is generated.

Quantitative results are presented in \tabrefede{tab:flowevluation}. We use the average endpoint error (AEE) and the percentage of points with endpoint error greater than 3 pixels to compare our FlowNet architectures against three EV-FlowNet from literature; two of them trained with frame- (FW) \cite{zhu2018ev} and event-warping (EW) \cite{zhu2019unsupervised} SSL proxy losses on MVSEC \cite{zhu2018multivehicle}, and one trained with synthetic ground-truth data (GT) \cite{stoffregen2020train}. For our networks, the number of input events per pixel was set to $0.3$. Error metrics were only acquired over pixels with valid ground-truth data and at least one event; and, for comparison, we used the quantitative results reported in \cite{zhu2019unsupervised,stoffregen2020train}.

From \tabrefede{tab:flowevluation}, the first noticeable aspect is the accuracy gap between EV-FlowNet$\text{}_{\text{GT-SIM}}$ and the rest of networks. Training with ground-truth dense optical flow entails certain ability to resolve the aperture problem \cite{de2020neural} that most SSL approaches lack. Regarding the latter, our EV-FlowNet performs consistently better than EV-FlowNet$\text{}_{\text{FW-MVSEC}}$ in all sequences except for outdoor\_day1, but underperforms EV-FlowNet$\text{}_{\text{EW-MVSEC}}$ despite using the same architecture and training procedure. We believe this is mostly due to the different training datasets and the fact that we did not fine-tune the number of input events for this evaluation. Further, note that these literature architectures were trained on a very similar driving sequence from MVSEC, while our training data is much more diverse in terms of optical flow vectors \cite{delmerico2019we}.

\begin{figure*}[!t]
	\centering
	\includegraphics[width=0.975
	\textwidth]{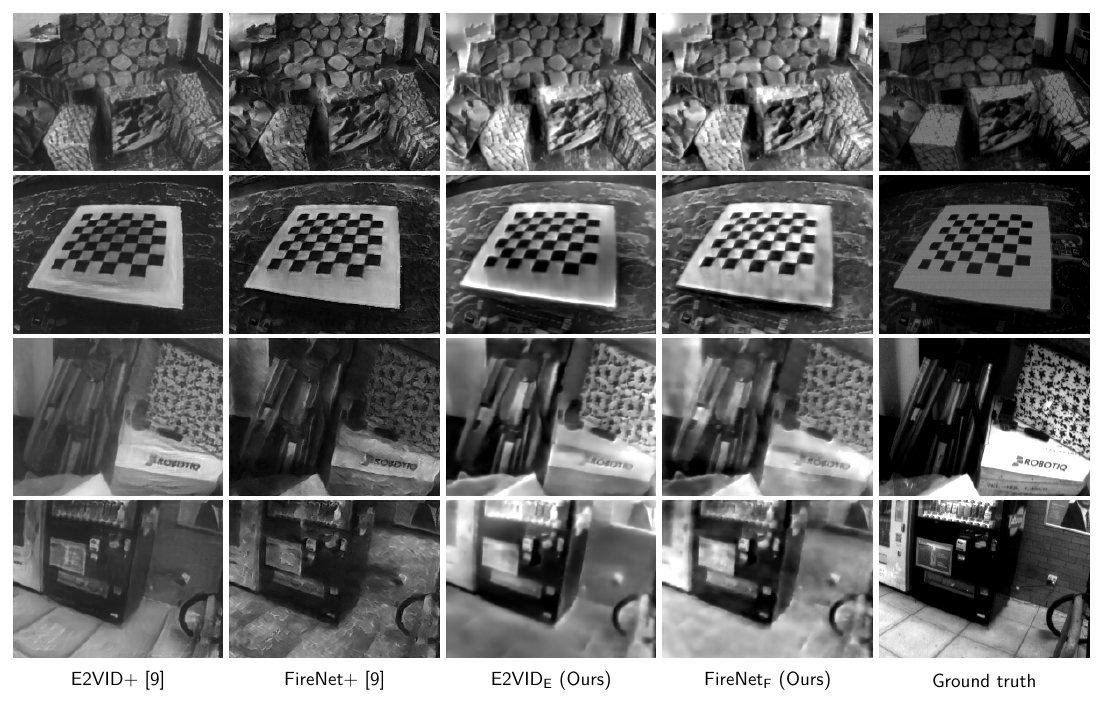}
	\caption{Qualitative comparison of our method with the state-of-the-art E2VID+ and FireNet+ architectures \cite{stoffregen2020train} on sequences from the ECD \cite{mueggler2017event} and HQF \cite{stoffregen2020train} datasets. Local histogram equalization not used for this comparison.}
	\label{fig:qualitativerecon}
\end{figure*}

Using our EV-FlowNet as reference, \tabrefede{tab:flowevluation} shows that the proposed FireFlowNet is characterized by a comparable accuracy despite the significant reduction in model complexity. This performance drop is likely due to the narrow receptive field of the architecture, which entails limitations due to the aperture problem. Regarding computational cost, \tabrefede{tab:flowmetrics} shows that FireFlowNet runs ${\sim}1.3$-$2.2$ times faster than EV-FlowNet on GPU, requiring less than ${\sim}30\%$ of FLOPS per forward-pass.

\begin{table}[!t]
	\caption{Quantitative evaluation of our FlowNet architectures on the ECD \cite{mueggler2017event} and HQF \cite{stoffregen2020train} datasets. For each dataset, we report the mean FWL \cite{stoffregen2020train} (higher is better, $\uparrow$). Best in bold, runner up underlined.
		%	The breakdown of the results can be found in the supplementary material.
	}
	\label{tab:flow}
	\centering
	\resizebox{0.75\linewidth}{!}{%
		{\renewcommand{\arraystretch}{1.1} 
			\begin{tabular}{lcc}
				\thickhline
				& \multicolumn{1}{c}{ECD$^*$}&\multicolumn{1}{c}{HQF}\\\hline
				EV-FlowNet$_\text{FW-MVSEC}$ \cite{zhu2018ev} & 1.36 & 1.25\\
				EV-FlowNet$_\text{GT-SIM}$ \cite{stoffregen2020train} & \textbf{1.51} & 1.39\\
				EV-FlowNet$_\text{EW-DR}$ (Ours) & 1.31 & \underline{1.51}\\
				FireFlowNet$_\text{EW-DR}$ (Ours) & \underline{1.39} & \textbf{1.58}\\
				\thickhline
				\multicolumn{3}{l}{\small $^*$Sequence cuts in the supplementary material.}
	\end{tabular}}}
\end{table}

For completeness, we also evaluate our FlowNet architectures on the ECD \cite{mueggler2017event} and HQF \cite{stoffregen2020train} datasets via the Flow Warp Loss (FWL) \cite{stoffregen2020train}. This metric, which does not require ground-truth data, measures the sharpness of the IWE in relation to that of the original partition of events. Similarly to \cite{stoffregen2020train}, we set the number of input events to 50k for all sequences in this evaluation\footnote{Note that the formulation of the FWL metric is sensitive to the number of input events \cite{stoffregen2020train}.}. \tabrefede{tab:flow} shows that both our FlowNet architectures, which are specifically trained to perform event deblurring (see \secrefede{sec:flow}), are in line with or outperform the state-of-the-art EV-FlowNet trained with either frames \cite{zhu2018ev} or synthetic ground truth \cite{stoffregen2020train} according to this metric. More interestingly, FireFlowNet outperforms our EV-FlowNet in both datasets. A qualitative evaluation of our FlowNet architectures can be found in the supplementary material.

\begin{figure*}[!t]
	\centering
	\includegraphics[width=0.985
	\textwidth]{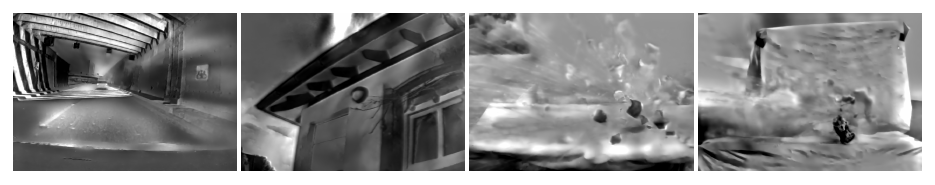}
	\caption{Qualitative results of our E2VID$_\text{E}$ on sequences from the High Speed and HDR Dataset \cite{rebecq2019high}.}
	\label{fig:rebecq}
\end{figure*}

\subsection{Reconstruction Evaluation}

We evaluated the accuracy of our ReconNet architectures against the DAVIS240C \cite{brandli2014240} frames from the ECD \cite{mueggler2017event} and HQF \cite{stoffregen2020train} datasets, and compared their performance to the state-of-the-art of image reconstruction networks trained with ground-truth supervision: E2VID \cite{rebecq2019high}, FireNet \cite{scheerlinck2020fast}, E2VID+ \cite{stoffregen2020train}, and FireNet+ \cite{stoffregen2020train}. Super resolution and adversarial methods are not considered in this comparison. We used the results and code provided by Stoffregen, Scheerlinck \textit{et al.}\ \cite{stoffregen2020train} for the quantitative and qualitative evaluations. The subscripts F and E indicate whether our networks were trained together with FireFlowNet or EV-FlowNet.

\begin{table}[!t]
	\caption{Quantitative evaluation of our ReconNet architectures on the ECD \cite{mueggler2017event} and HQF \cite{stoffregen2020train} datasets. For each dataset, we report the mean MSE ($\downarrow$), SSIM \cite{wang2004image} ($\uparrow$) and LPIPS \cite{zhang2018unreasonable} ($\downarrow$). Best in bold; runner up underlined. 
		%Breakdown of the results in the supplementary material.
	}
	\label{tab:reconstruction}
	\centering
	\resizebox{\linewidth}{!}{%
		{\renewcommand{\arraystretch}{1.1} 
			\begin{tabular}{lccccccc}
				\thickhline
				& \multicolumn{3}{c}{ECD$^*$}& &\multicolumn{3}{c}{HQF}\\\cline{2-4}\cline{6-8}
				& MSE & SSIM & LPIPS && MSE & SSIM & LPIPS \\\hline
				E2VID \cite{rebecq2019high} & 0.08 & 0.54 & 0.37 && 0.14 & 0.46 & 0.45 \\
				FireNet \cite{rebecq2019high} & 0.06 & \underline{0.57} & \underline{0.29} && 0.07 & \underline{0.48} & 0.42 \\
				E2VID+ \cite{stoffregen2020train} & \textbf{0.04} & \textbf{0.60} & \textbf{0.27} && \textbf{0.03} & \textbf{0.57} & \textbf{0.26} \\
				FireNet+ \cite{stoffregen2020train} & \underline{0.06} & 0.51 & 0.32 && \underline{0.05} & 0.47 & \underline{0.36} \\
				E2VID$\text{}_{\text{F}}$ (Ours)  & 0.07 & 0.52 & 0.38 && 0.07 & 0.44 & 0.47 \\
				E2VID$\text{}_{\text{E}}$ (Ours)  & 0.06 & 0.55 & 0.37 && 0.06 & 0.48 & 0.47 \\
				FireNet$\text{}_{\text{F}}$ (Ours)  & 0.06 & 0.52 & 0.38 && 0.06 & 0.46 & 0.47 \\
				FireNet$\text{}_{\text{E}}$ (Ours)  & 0.06 & 0.51 & 0.41 && 0.06 & 0.46 & 0.51 \\
				\thickhline
				\multicolumn{8}{l}{\small $^*$Sequence cuts in the supplementary material.}
	\end{tabular}}}
\end{table}

\begin{figure}[!t]
	\centering
	\includegraphics[width=0.475
	\textwidth]{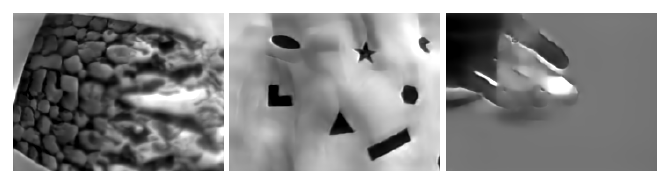}
	\caption{Common failure cases of our SSL framework, namely motion blur in case of suboptimal optical flow estimation (left), ghosting artifacts in large texture-less regions (center), and inconsistent reconstructions due to the lack of information about the initial brightness $L_0$ (right).}
	\label{fig:limitations}
\end{figure}

For all methods, reconstructions were generated at each DAVIS frame timestamp. We first applied local histogram equalization \cite{yadav2014contrast} to both frames, and then computed mean squared error (MSE), structural similarity (SSIM) \cite{wang2004image}, and perceptual similarity (LPIPS) \cite{zhang2018unreasonable}. Only for this evaluation, instead of using a fixed number of input events, we used all the events \textit{in between DAVIS frames}, thus generating image sets with the same number of frames as the ground truth. Quantitative results are presented in \tabrefede{tab:reconstruction}, and are supported by qualitative results in \figrefedetwo{fig:qualitativerecon}{fig:rebecq}. Additional results can be found in the supplementary material.

%we reconstructed a sets of images for each sequence with the same number of frames as the ground truth by encoding all the events in \textit{between reconstructed frames} into input voxel grids. 

Despite not using any ground-truth data during training, results show that our method is in line with the state-of-the-art in terms of reconstruction accuracy. Quantitatively, the error metrics of all our ReconNet architectures closely resemble the results obtained with the original E2VID and FireNet, but the accuracy gap increases if compared against these same networks trained with the refined data augmentation mechanisms from Stoffregen, Scheerlinck \textit{et al.}\ \cite{stoffregen2020train}. This gap is particularly notable in the LPIPS loss because these literature networks are specifically trained to maximize perceptual similarity to ground-truth frames. On the other hand, there is no major quantitative difference between the evaluated versions of ReconNet, regardless of their architecture or the accompanying flow network.

Qualitative results confirm that our method reconstructs high quality HDR images. However, it is possible to identify several differences with respect to the state-of-the-art. Firstly, our images appear less sharp. Our architectures learn to correlate the spatial gradients of the estimated brightness $\hat{L}$ to the averaged IWE (see \secrefede{sec:recons}). This entails that the reconstructed images are affected by the accuracy of the optical flow. Suboptimal optical flow estimations lead to imperfect event deblurring during training, which in turn is reflected in the reconstructed images as motion blur. Note that this blur diminishes when using an appropriate fixed number of input events for each sequence. Secondly, the dynamic range of the images differs. State-of-the-art methods learn to map the input events into bounded estimates of $\hat{L}$ 
%(determined by sigmoid activations in the last layers) 
via supervised learning. On the contrary, our brightness estimate is unbounded, and normalization is used to encode this signal as bounded images. Besides this, there is no significant difference between the evaluated ReconNet versions, despite the limited smoothing capabilities of FireNet. Lastly, although our method does not suffer from the stretch marks mostly present in FireNet+ images, it is characterized by three common failure cases. As shown in \figrefede{fig:limitations}, these are: (i) the aforementioned motion blur, (ii) ``ghosting'' artifacts in large texture-less regions due to limited extrapolation of edge information, and (iii) incoherent reconstructions due to the lack of information about the initial brightness $L_0$.

\section{Conclusion}
In this paper, we went back to the basics of event cameras and presented the first self-supervised learning-based approach to event-based image reconstruction, which does not rely on any ground-truth or synthetic data during training. Instead, our SSL method makes use of the flow-intensity relation used by early methods to reconstruct the frames that best satisfy the input events and the estimated optical flow. Results confirm that our method performs almost as well as the state-of-the-art, but that the reconstructed images are characterized by several artifacts that need to be addressed by future work. Additionally, we presented FireFlowNet: a fast, lightweight neural network that performs event-based optical flow estimation. We believe this work shows the exciting potential of SSL to take over the research on image reconstruction from event data, and it opens up avenues for further improvement by leveraging the great amount of unlabeled event data available. Moreover, we have proposed a general self-supervised learning framework that can be extended in multiple ways via more sophisticated reconstruction losses and other event-based optical flow algorithms.

{\small
	\bibliographystyle{IEEEtran}
	\input{cvpr.bbl}

}

\clearpage
\appendix

\section{Sequence Cuts}\label{appendix:cuts}

The DAVIS frames accompanying the frequently used Event-Camera Dataset \cite{mueggler2017event} usually suffer from motion blur and under/overexposure. For this reason, we only evaluate reconstruction accuracy on sections of this dataset in which the frames appear to be of high quality. The exact cut times are adopted from \cite{stoffregen2020train} and shown in \tabrefede{tab:cuts}. Additionally, we only evaluate optical flow accuracy on these sections to remain comparable to the results reported in \cite{stoffregen2020train}.

\begin{table}[!h]
	\caption{Sequence cuts used for evaluation on the Event-Camera Dataset \cite{mueggler2017event}. Adopted from \cite{stoffregen2020train}.
	}
	\label{tab:cuts}
	\centering
	\resizebox{0.7\linewidth}{!}{%
		{\renewcommand{\arraystretch}{1.1} 
			\begin{tabular}{ccc}
				\thickhline
				Sequence & Start [s] & End [s]\\\hline
				boxes\_6dof\_cut & $5.0$ & $20.0$\\
				calibration\_cut & $5.0$ & $20.0$\\
				dynamic\_6dof\_cut & $5.0$ & $20.0$\\
				office\_zigzag\_cut & $5.0$ & $12.0$\\
				poster\_6dof\_cut & $5.0$ & $20.0$\\
				shapes\_6dof\_cut & $5.0$ & $20.0$\\
				slider\_depth\_cut & $1.0$ & $2.5$\\
				\thickhline
	\end{tabular}}}
\end{table}

\section{Impact of Event Deblurring}

As discussed in this work, our self-supervised image reconstruction framework is designed around the event-based photometric constancy equation. While the right-hand side of this equation is obtained via the dot product between the warped spatial gradients of the last reconstructed image and the estimated optical flow; we propose that the left-hand side is obtained by integrating the deblurred (and averaged) input events. Since the main supervisory signal used to train our image reconstruction architectures comes from the comparison of the two sides of this equation, after training, the spatial gradients of the reconstructed images are correlated with the integrated events. These events, if not warped to the timestamp of the reconstructed frame, would introduce motion blur into the images. The amount of motion blur would depend on the density of events and on the length of the partition of events.

\begin{figure}[!t]
	\centering
	\includegraphics[width=0.4736
	\textwidth]{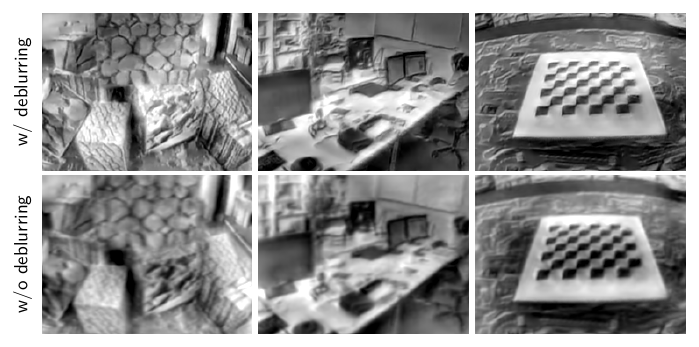}
	\caption{Qualitative evaluation of the impact of event deblurring on the quality of the reconstructed frames on sequences from the ECD \cite{mueggler2017event} dataset.}
	\label{fig:fede}
\end{figure}

To validate this approach, we conducted an ablation study in which we trained the same ReconNet architecture (accompanied by the same pre-trained optical flow network) with and without event deblurring prior to event integration. Quantitative results are presented in \tabrefede{tab:reconstruction}, and are supported by qualitative results in \figrefede{fig:fede}. As shown, event deblurring is a crucial mechanism to reconstruct sharp images from the events. Without it, the reconstructed frames appear less sharp for the same number of input events, and the network is characterized by significantly worse error metrics on the evaluation datasets.

\begin{table}[!t]
	\caption{Quantitative evaluation of the impact of event deblurring prior to event integration on the ECD \cite{mueggler2017event} and HQF \cite{stoffregen2020train} datasets. For each dataset, we report the mean MSE ($\downarrow$), SSIM \cite{wang2004image} ($\uparrow$) and LPIPS \cite{zhang2018unreasonable} ($\downarrow$). Best in bold. 
		%Breakdown of the results in the supplementary material.
	}
	\label{tab:reconstruction}
	\centering
	\resizebox{\linewidth}{!}{%
		{\renewcommand{\arraystretch}{1.1} 
			\begin{tabular}{lccccccc}
				\thickhline
				& \multicolumn{3}{c}{ECD$^*$}& &\multicolumn{3}{c}{HQF}\\\cline{2-4}\cline{6-8}
				& MSE & SSIM & LPIPS && MSE & SSIM & LPIPS \\\hline
				E2VID$\text{}_{\text{E}}$ (w/ deblurring)  & \textbf{0.06} & \textbf{0.55} & \textbf{0.37} && \textbf{0.06} & \textbf{0.48} & \textbf{0.47} \\
				E2VID$\text{}_{\text{E}}$ (w/o deblurring)  & 0.14 & 0.30 & 0.58  && 0.11 & 0.28 & 0.64 \\
				\thickhline
				\multicolumn{8}{l}{\small $^*$Sequence cuts in \tabrefede{tab:cuts}.}
	\end{tabular}}}
\end{table}

\section{Additional Quantitative Results}\label{appendix:breakdown}

A breakdown of the quantitative results of our FlowNet and ReconNet architectures on the ECD \cite{mueggler2017event} and HQF \cite{stoffregen2020train} datasets can be found in \tabrefedetwo{tab:reconstructionseparatedflow}{tab:reconstructionseparated}, respectively.

\section{Additional Qualitative Results}\label{appendix:qualitative}

\begin{figure}[!t]
	\centering
	\includegraphics[width=0.2
	\textwidth]{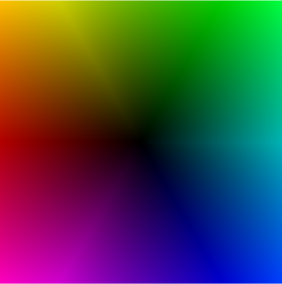}
	\caption{Optical flow field color-coding scheme. Direction is encoded incolor hue, and speed in color brightness. }
	\label{fig:flowcode}
\end{figure}

\begin{figure*}[!t]
	\centering
	\begin{subfigure}[t]{0.5\textwidth}
		\centering
		\includegraphics[width=1
		\textwidth]{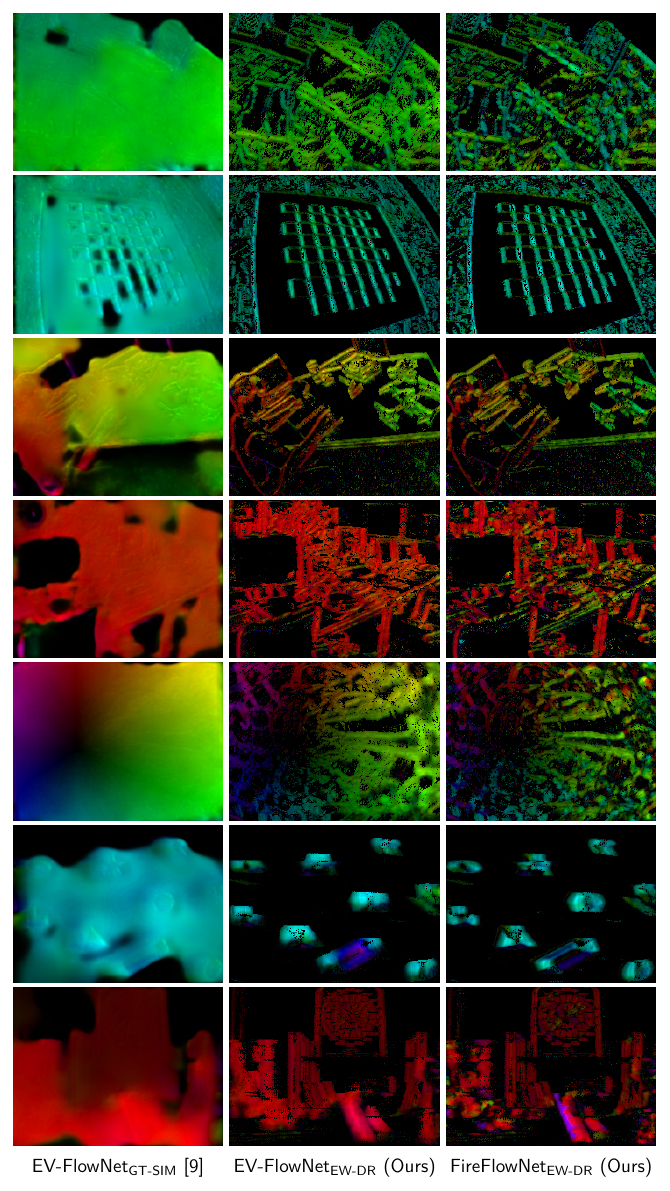}
		\caption{ECD dataset.}
	\end{subfigure}%
	\begin{subfigure}[t]{0.5\textwidth}
		\centering
		\includegraphics[width=1
		\textwidth]{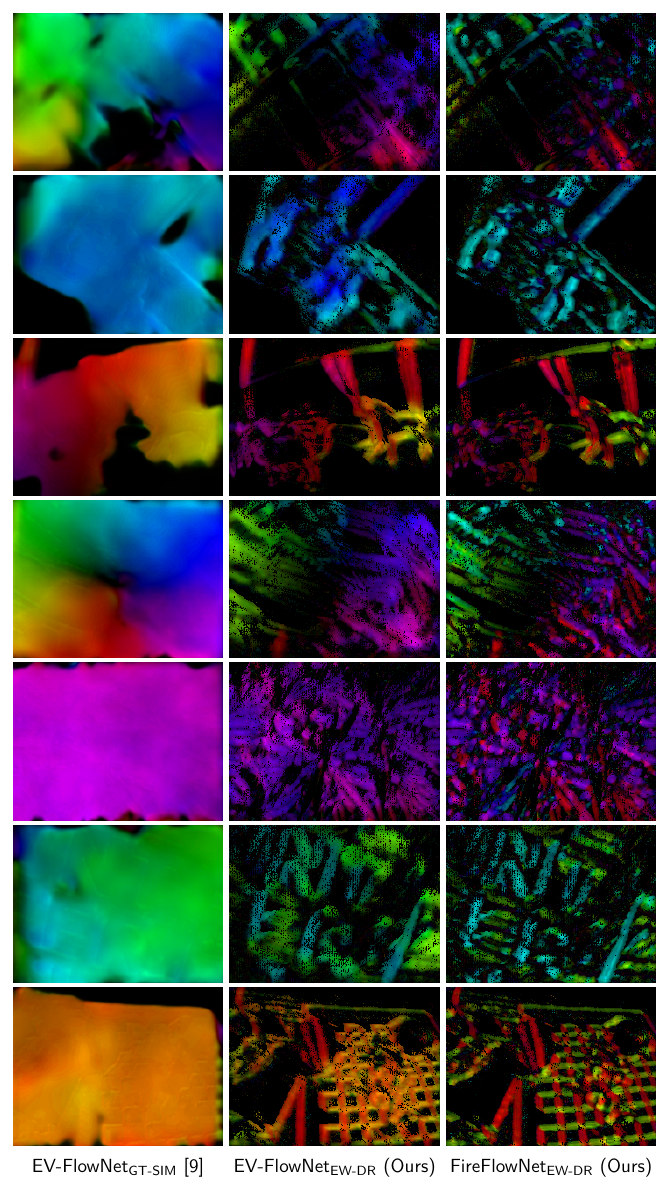}
		\caption{HQF dataset.}
	\end{subfigure}%
	\caption{Additional qualitative comparison of our FlowNet architectures with the state-of-the-art EV-FlowNet \cite{stoffregen2020train} on sequences from the ECD \cite{mueggler2017event} and HQF \cite{stoffregen2020train} dataset.}
	\label{fig:addqual4}
\end{figure*}

\begin{figure*}[!t]
	\centering
	\includegraphics[width=1
	\textwidth]{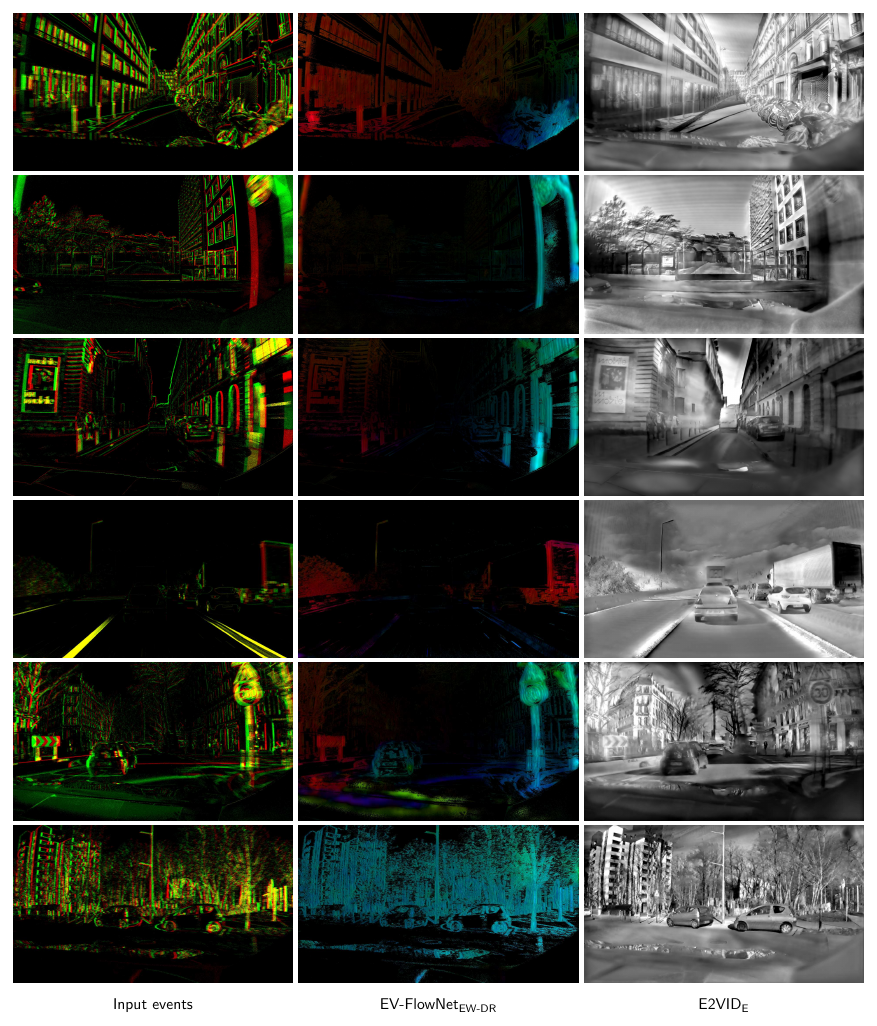}
	\caption{Additional qualitative results on sequences from Prophesee's high-resolution automotive dataset \cite{perot2020learning}. }
	\label{fig:prophesee}
\end{figure*}

\figrefedethree{fig:addqual4}{fig:addqual1}{fig:addqual3} show additional qualitative results of our FlowNet and ReconNet architectures on the ECD \cite{mueggler2017event} and HQF \cite{stoffregen2020train} datasets. Lastly, \figrefede{fig:prophesee} shows qualitative results on the high-resolution automotive dataset recently released by Prophesee \cite{perot2020learning}. The optical flow color-coding scheme for \figrefedetwo{fig:addqual4}{fig:prophesee} can be found in \figrefede{fig:flowcode}.

\begin{table*}[!t]
	\caption{Breakdown of the quantitative evaluation of our FlowNet architectures on the ECD \cite{mueggler2017event} and HQF \cite{stoffregen2020train} datasets. For each dataset, we report the FWL \cite{stoffregen2020train} ($\uparrow$).}
	\label{tab:reconstructionseparatedflow}
	\centering
	\resizebox{0.915\linewidth}{!}{%
		{\renewcommand{\arraystretch}{1.1} 
			\begin{tabular}{lcccc}
				\thickhline
				& EV-FlowNet$_\text{FW-MVSEC}$ \cite{zhu2018ev} & EV-FlowNet$_\text{GT-SIM}$ \cite{stoffregen2020train} & EV-FlowNet$_\text{EW-DR}$ (Ours) & FireFlowNet$_\text{EW-DR}$ (Ours) \\\hhline{=====}
				& \multicolumn{4}{c}{\textbf{ECD$^*$}}\\
				boxes\_6dof\_cut & 1.42 & 1.46 & 1.22 & 1.37\\
				calibration\_cut & 1.20 & 1.31 & 1.11 & 1.22\\
				dynamic\_6dof\_cut & 1.37 & 1.39 & 1.22 & 1.33\\
				office\_zigzag\_cut & 1.13 & 1.11 & 1.09 & 1.18\\
				poster\_6dof\_cut & 1.50 & 1.56 & 1.20 & 1.34\\
				shapes\_6dof\_cut & 1.15 & 1.57  & 1.51 & 1.38\\
				slider\_depth\_cut & 1.73 & 2.17 & 1.80 & 1.88\\\hline
				\textbf{Mean} & \textbf{1.36} & \textbf{1.51} & \textbf{1.31} & \textbf{1.39}\\\hhline{=====}
				& \multicolumn{4}{c}{\textbf{HQF}}\\
				bike\_day\_hdr & 1.22 & 1.23 & 1.49 & 1.52\\
				boxes & 1.75 & 1.80 & 1.68 & 1.72\\
				desk & 1.23 & 1.35 & 1.35 & 1.42\\
				desk\_fast & 1.43 & 1.50 & 1.42 & 1.47\\
				desk\_hand\_only & 0.95 & 0.85 & 1.14 & 1.23\\
				desk\_slow & 1.01 & 1.08 & 1.23 & 1.27\\
				engineering\_posters & 1.50 & 1.65 & 1.65 & 1.71\\
				high\_texture\_plants & 0.13 & 1.68 & 1.71 & 1.77\\
				poster\_pillar\_1 & 1.20 & 1.24 & 1.39 & 1.45\\
				poster\_pillar\_2 & 1.16 & 0.96 & 1.10 & 1.18\\
				reflective\_materials & 1.45 & 1.57 & 1.62 & 1.63\\
				slow\_and\_fast\_desk & 0.93 & 0.99 & 1.68 & 1.77\\
				slow\_hand & 1.64 & 1.56 & 1.90 & 1.96\\
				still\_life & 1.93 & 1.98 & 1.76 & 1.97\\\hline
				\textbf{Mean} & \textbf{1.25} & \textbf{1.39} & \textbf{1.51} & \textbf{1.58}\\
				\thickhline
	\end{tabular}}}
\end{table*}

\begin{table*}[!t]
	\caption{Breakdown of the quantitative results of our ReconNet architectures on the ECD \cite{mueggler2017event} and HQF \cite{stoffregen2020train} datasets. For each sequence, we report the MSE ($\downarrow$), SSIM \cite{wang2004image} ($\uparrow$) and LPIPS \cite{zhang2018unreasonable} ($\downarrow$). The F and E subscripts determine whether our networks were trained in combination with FireFlowNet or EV-FlowNet, respectively.}
	\label{tab:reconstructionseparated}
	\centering
	\resizebox{1\linewidth}{!}{%
		{\renewcommand{\arraystretch}{1.1} 
			\begin{tabular}{lcccccccccccccc}
				\thickhline
				& \multicolumn{4}{c}{MSE}&& \multicolumn{4}{c}{SSIM}&& \multicolumn{4}{c}{LPIPS}\\\cline{2-5}\cline{7-10}\cline{12-15}
				& FireNet$\text{}_{\text{F}}$ & FireNet$\text{}_{\text{E}}$ & E2VID$\text{}_{\text{F}}$ & E2VID$\text{}_{\text{E}}$ && FireNet$\text{}_{\text{F}}$ & FireNet$\text{}_{\text{E}}$ & E2VID$\text{}_{\text{F}}$ & E2VID$\text{}_{\text{E}}$ && FireNet$\text{}_{\text{F}}$ & FireNet$\text{}_{\text{E}}$ & E2VID$\text{}_{\text{F}}$ & E2VID$\text{}_{\text{E}}$ \\\hhline{===============}
				& \multicolumn{14}{c}{\textbf{ECD$^*$}}\\
				boxes\_6dof\_cut & 0.0533 & 0.0554 & 0.0540 & 0.0541 && 0.5705 & 0.5538 & 0.5785 & 0.5997&& 0.3736 & 0.4170 & 0.3776 & 0.3781 \\
				calibration\_cut & 0.0531 & 0.0620 & 0.0779 & 0.0677 && 0.5464 & 0.5356 & 0.5445 & 0.5594&& 0.2770 & 0.3046 & 0.2982 & 0.2937 \\
				dynamic\_6dof\_cut & 0.0950 & 0.0780 & 0.1030 & 0.0845 && 0.4037 & 0.4036 & 0.4123 & 0.4519&& 0.4773 & 0.4969 & 0.4576 & 0.4424 \\
				office\_zigzag\_cut & 0.0452 & 0.0427 & 0.0442 & 0.0617 && 0.5019 & 0.5033 & 0.4970 & 0.4807&& 0.3634 & 0.4122 & 0.3350 & 0.3485 \\
				poster\_6dof\_cut & 0.0592 & 0.0567 & 0.0593 & 0.0521 && 0.5385 & 0.5211 & 0.5613 & 0.5823&& 0.4039 & 0.4396 & 0.3941 & 0.3909 \\
				shapes\_6dof\_cut & 0.0500 & 0.0928 & 0.0608 & 0.0594 && 0.5719 & 0.5262 & 0.5673 & 0.6297&& 0.4303 & 0.4313 & 0.4532 & 0.3554 \\
				slider\_depth\_cut & 0.0612 & 0.0613 & 0.0840 & 0.0660 && 0.5200 & 0.5265 & 0.4758 & 0.5174&& 0.3613 & 0.3834 & 0.3536 & 0.3728 \\\hline
				\textbf{Mean} & \textbf{0.0595} & \textbf{0.0641} & \textbf{0.0690} & \textbf{0.0636} && \textbf{0.5218} & \textbf{0.5100} & \textbf{0.5195} & \textbf{0.5459}&& \textbf{0.3838} & \textbf{0.4121} & \textbf{0.3813} & \textbf{0.3688} \\\hhline{===============}
				& \multicolumn{14}{c}{\textbf{HQF}}\\
				bike\_day\_hdr & 0.0629 & 0.0587 & 0.0552 & 0.0519 && 0.4317 & 0.4471 & 0.4574 &  0.4835&& 0.5248 & 0.5584 & 0.5028 & 0.5266 \\
				boxes & 0.0596 & 0.0549 & 0.0694 & 0.0562 && 0.4885 & 0.4912 & 0.4853 & 0.5190&& 0.3994 & 0.4439& 0.4108 & 0.4164 \\
				desk & 0.0619 & 0.0649 &  0.0817 & 0.0697 && 0.4776 & 0.4779 & 0.4677 & 0.4972&& 0.3938 & 0.4373 & 0.4018 & 0.3914 \\
				desk\_fast & 0.0588 & 0.0624 & 0.0711 & 0.0637 && 0.4935 & 0.4882 & 0.5027 & 0.5238&& 0.4482 & 0.4999 & 0.4425 & 0.4515 \\
				desk\_hand\_only & 0.0805 & 0.0910 & 0.0755 & 0.0594 && 0.5143 & 0.5106 & 0.5134 & 0.5545&& 0.5971 & 0.6202 & 0.5619 & 0.5438 \\
				desk\_slow & 0.0783 & 0.0894 & 0.0976 & 0.0759 && 0.5011 & 0.4341 & 0.2852 & 0.4998&& 0.5214 & 0.6029 & 0.6689 & 0.5253 \\
				engineering\_posters & 0.0570 & 0.0541 & 0.0783 & 0.0656 && 0.4690 & 0.4776 & 0.4456 & 0.4797&& 0.4250 & 0.4417& 0.4345 & 0.4528 \\
				high\_texture\_plants & 0.0579 & 0.0581 & 0.0687 & 0.0653 && 0.4689 & 0.4705 & 0.4081 & 0.4404&& 0.3618 & 0.4054 & 0.3895 & 0.3825 \\
				poster\_pillar\_1 & 0.0653 & 0.0623 & 0.0726 & 0.0641 && 0.3132 & 0.3121 & 0.3340 & 0.3455&& 0.5532 & 0.5720 & 0.5144 & 0.5455 \\
				poster\_pillar\_2 & 0.0638 & 0.0605 & 0.0644 & 0.0532 && 0.3569 & 0.3814 & 0.3881 & 0.4119&& 0.5968 & 0.6059 & 0.5643 & 0.5737 \\
				reflective\_materials & 0.0506 & 0.0517 & 0.0566 & 0.0528 && 0.4621 & 0.4705 & 0.4779 & 0.5032&& 0.4235 & 0.4655 & 0.4254 & 0.4493 \\
				slow\_and\_fast\_desk & 0.0701 & 0.0648 & 0.0620 & 0.0699 && 0.4503 & 0.4584 & 0.4805 & 0.4850&& 0.4565 & 0.4903 & 0.4200 & 0.4321 \\
				slow\_hand & 0.0824 & 0.0667 & 0.0736 & 0.0614 && 0.4123 & 0.4246 & 0.4380 & 0.4647&& 0.5480 & 0.5651 & 0.4694 & 0.4937 \\
				still\_life & 0.0429 & 0.0419 & 0.0486 & 0.0469 && 0.5434 & 0.5413 & 0.5376 & 0.5470&& 0.3924 & 0.4400 & 0.4187 & 0.4515 \\\hline
				\textbf{Mean} & \textbf{0.0637} & \textbf{0.0629} & \textbf{0.0696} & \textbf{0.0611} && \textbf{0.4559} & \textbf{0.4561} & \textbf{0.4444} & \textbf{0.4825}&& \textbf{0.4744} & \textbf{0.5106} & \textbf{0.4732} & \textbf{0.4740} \\
				\thickhline
	\end{tabular}}}
\end{table*}

\begin{figure*}[!t]
	\centering
	\includegraphics[width=1
	\textwidth]{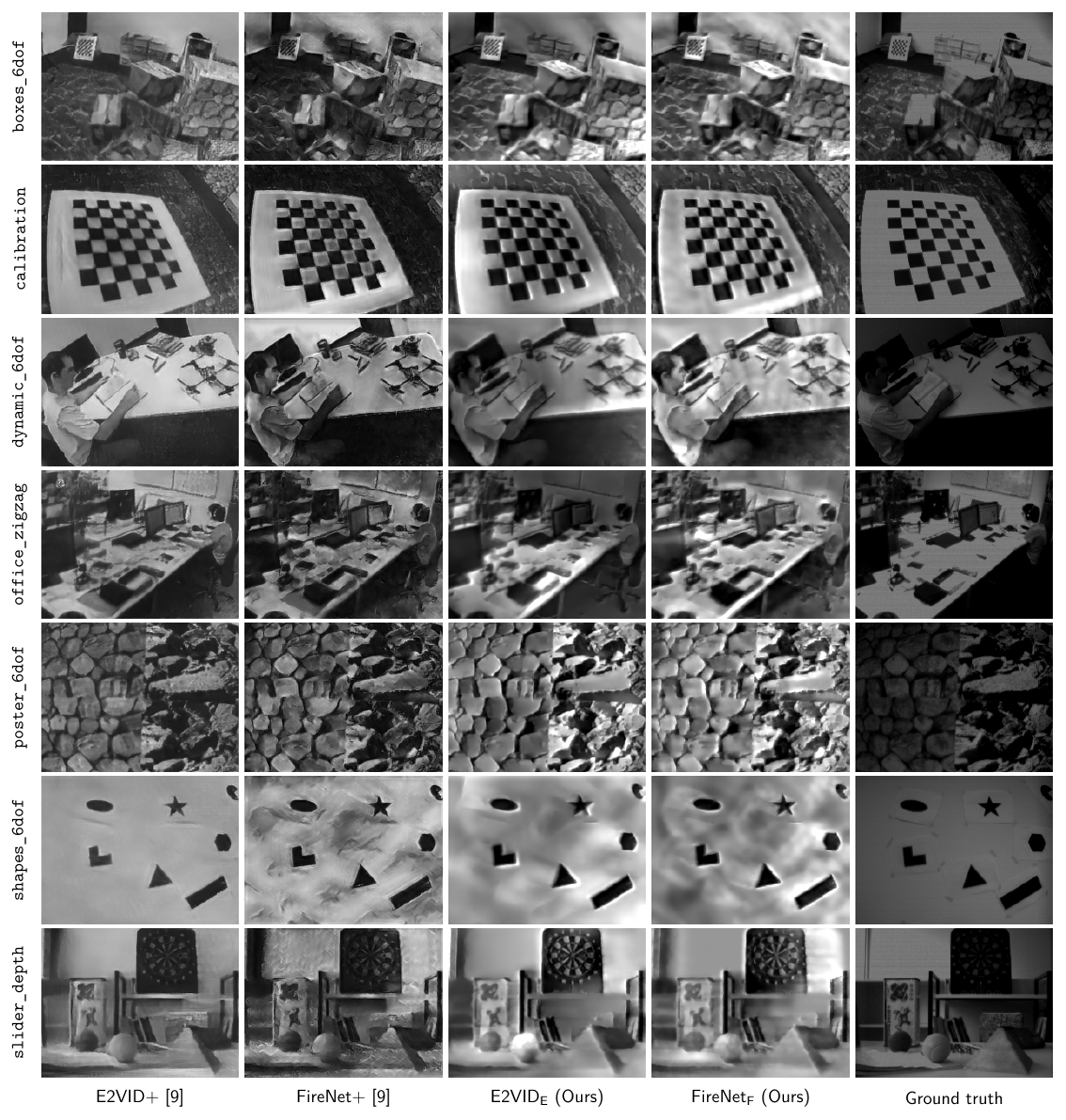}
	\caption{Additional qualitative comparison of our ReconNet architectures with the state-of-the-art E2VID+ and FireNet+ \cite{stoffregen2020train} on sequences from the ECD \cite{mueggler2017event} dataset. Local histogram equalization not used for this comparison.}
	\label{fig:addqual1}
\end{figure*}

\begin{figure*}[!t]
	\centering
	\includegraphics[width=1
	\textwidth]{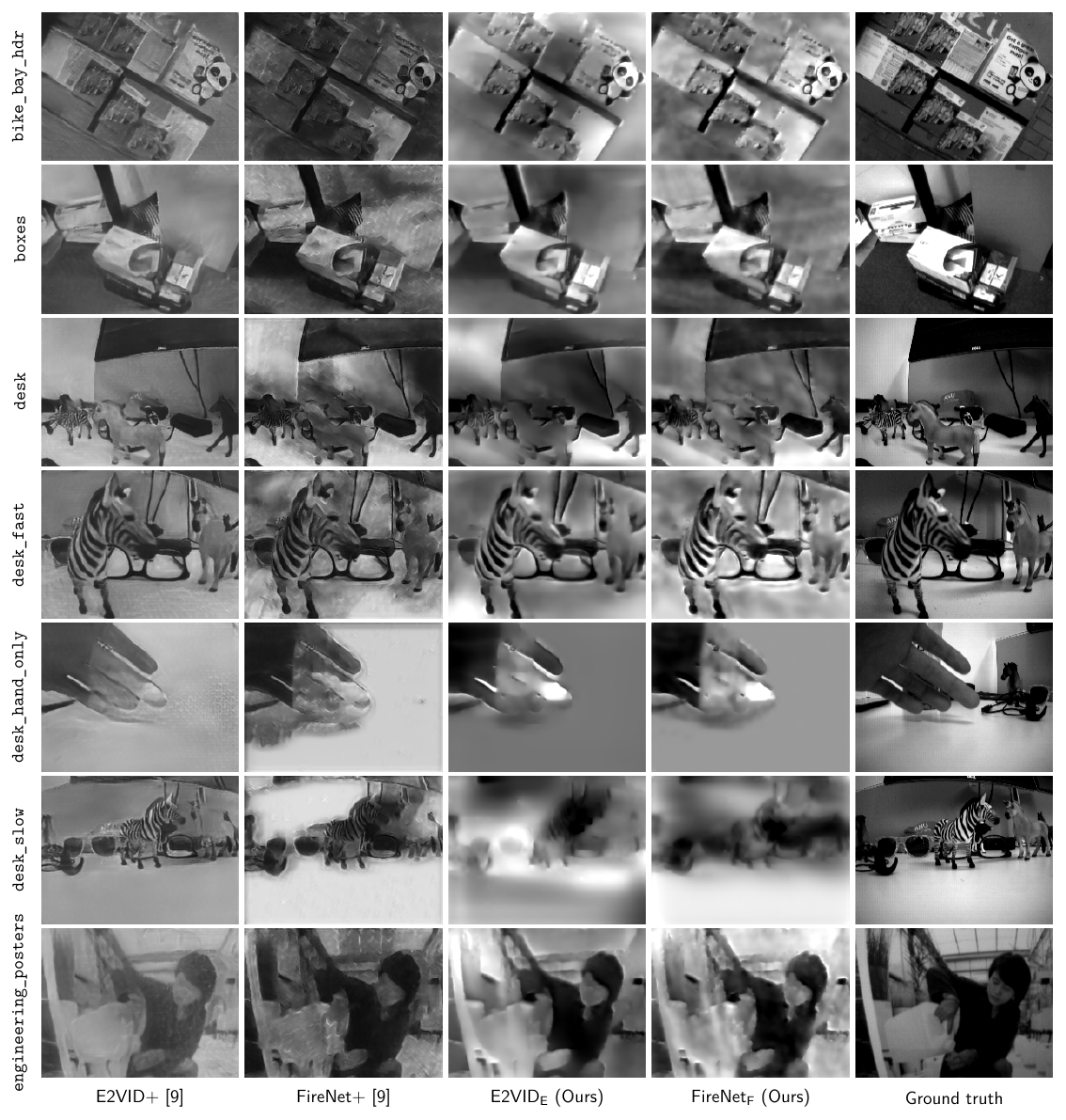}
	\label{fig:addqual2}
\end{figure*}

\begin{figure*}[!t]
	\centering
	\includegraphics[width=1
	\textwidth]{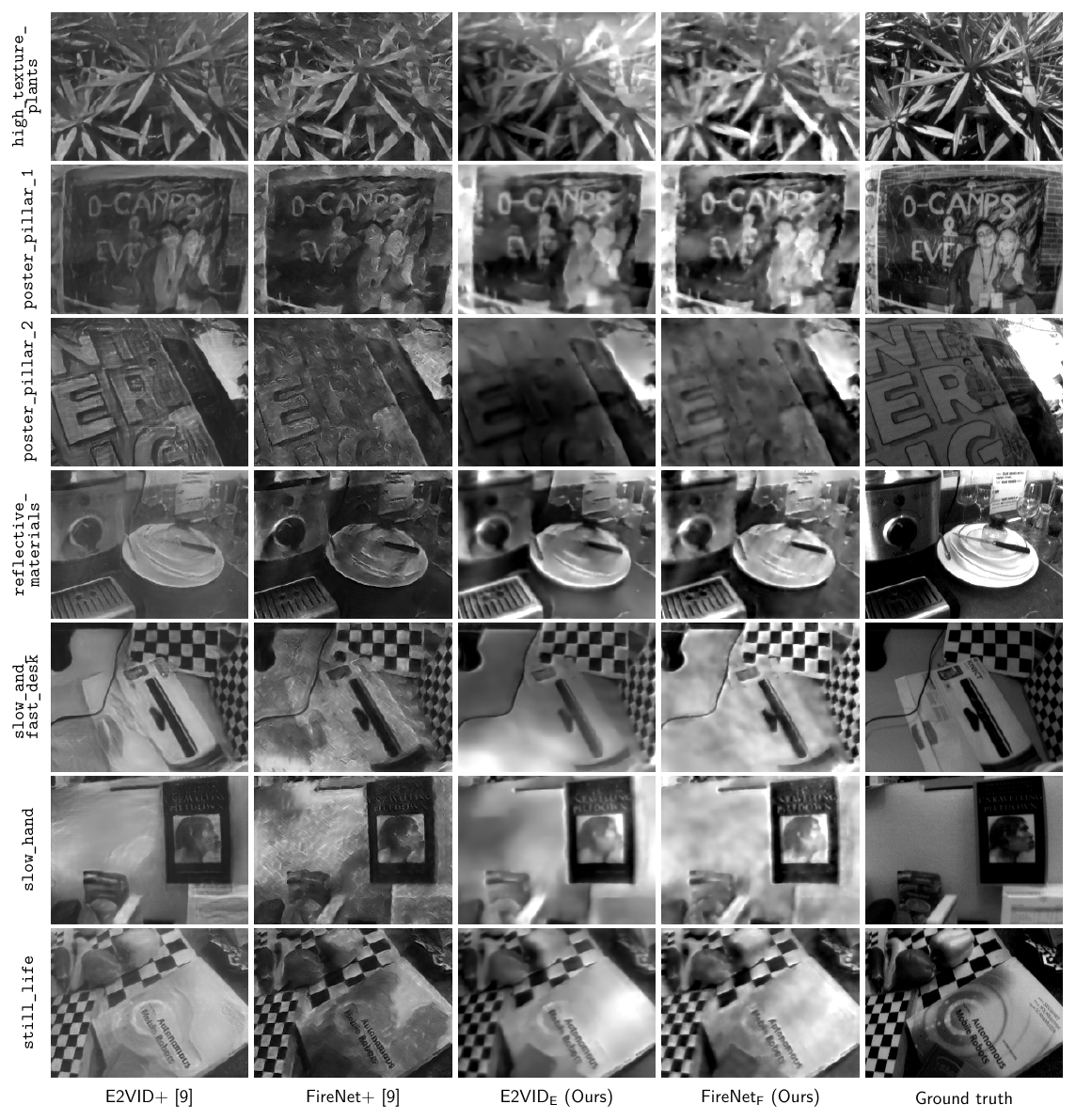}
	\caption{Additional qualitative comparison of our ReconNet architectures with the state-of-the-art E2VID+ and FireNet+ \cite{stoffregen2020train} on sequences from the HQF \cite{stoffregen2020train} dataset. Local histogram equalization not used for this comparison.}
	\label{fig:addqual3}
\end{figure*}

\end{document}